\documentclass{article}

\PassOptionsToPackage{numbers, compress}{natbib}
\usepackage[preprint]{neurips_2026}

\usepackage[utf8]{inputenc}
\usepackage[T1]{fontenc}
\usepackage{hyperref}
\usepackage{url}
\usepackage{booktabs}
\usepackage{amsfonts}
\usepackage{amsmath}
\usepackage{nicefrac}
\usepackage{microtype}
\usepackage{xcolor}

\usepackage{graphicx}
\usepackage{multirow}
\usepackage{algorithm}
\usepackage{algpseudocode}

\title{StableVQ: Practical Guidelines for Stable Vector-Quantized Tokenizer Training}

\makeatletter
\renewcommand{\@notice}{}
\makeatother
\author{%
  \normalfont
  \textbf{Bao Tang\textsuperscript{1,2,\S}}\quad
  \textbf{Jiahao Guo\textsuperscript{1,2,\S}}\quad
  \textbf{Haoxiang Cao\textsuperscript{2,3,\S}}\\
  \textbf{Wenyu Liu\textsuperscript{1}}\quad
  \textbf{Changqian Yu\textsuperscript{2,\textdagger}}\quad
  \textbf{Kun Gai\textsuperscript{2}}\quad
  \textbf{Xinggang Wang\textsuperscript{1,\textdagger}}\\[0.5em]
  \textsuperscript{1}Huazhong University of Science and Technology\quad
  \textsuperscript{2}KlingAI Research\\
  \textsuperscript{3}South China Normal University
}

\begin{document}

\maketitle

\begingroup
\renewcommand{\thefootnote}{\fnsymbol{footnote}}
\footnotetext[4]{Work done during internship at KlingAI Research.}
\footnotetext[2]{Corresponding authors: X. Wang (\texttt{xgwang@hust.edu.cn}) \& C. Yu (\texttt{y-changqian@outlook.com}).}
\endgroup
\vspace{-10pt}

\begin{abstract}
Vector Quantization (VQ) is fundamental to discrete visual tokenizers that power modern autoregressive and masked image generation models. While recent shared-projection codebook methods have substantially advanced codebook utilization, training stability remains a critical and underexplored challenge. We argue that the root cause lies in the \emph{entanglement} of the Encoder--Decoder and Codebook training: because neither module can reliably fulfill its own responsibility in isolation, the system can only function when the two subsystems happen to cooperate---a fragile condition that breaks down precisely when training is most stressed. We propose \textbf{StableVQ}, which revisits the proper learning objective of each module and resolves the problems that arise when each is trained to fulfill its own role independently. Concretely, (1) \textbf{Dynamic STE} corrects the instability in the Encoder's learning objective, enabling it to robustly optimize the reconstruction space under discrete regularization even when codebook utilization is low. (2) \textbf{Region VQ Loss} reconceives the Codebook's learning objective so that it can independently guarantee full tracking of the encoder output distribution, without relying on encoder oscillations to drive activation. (3) \textbf{Decoupled Schedule} recognizes that the distinct responsibilities of the Encoder--Decoder and the Codebook demand distinct optimization dynamics, and assigns each an independent learning rate schedule to ensure robust system-level behavior. Built on top of shared-projection codebooks, StableVQ is lightweight and introduces no learnable parameters. Experiments on ImageNet demonstrate consistent improvements in training stability, codebook utilization, and reconstruction quality across diverse codebook sizes and initialization settings.

\end{abstract}

\section{Introduction}
\label{sec:intro}

Visual tokenization has become a foundational component of modern generative vision systems~\cite{vqvae,vqgan,llamagen,var}. By mapping continuous image features to sequences of discrete tokens via a learned codebook, VQ-VAEs~\cite{vqvae} enable autoregressive transformers~\cite{llamagen,var,lfq}, masked generative models~\cite{maskgit}, and multimodal language models to operate over compact, structured visual representations. The expressiveness of the resulting vocabulary directly determines the upper bound on downstream generation quality.

A persistent obstacle in VQ training is \textbf{codebook collapse}, where most code vectors are never assigned, severely underutilizing model capacity. Recent \emph{shared-projection} methods~\cite{straighten,simvq,fvq} address this by reparameterizing codebook entries as $\tilde{\mathbf{e}}_k = f_\theta(\mathbf{e}_k)$ via a shared differentiable function, so that gradients propagate across the entire code distribution. This reframes VQ training as a \emph{distribution alignment} problem between the projected code distribution and the encoder output distribution, substantially advancing codebook utilization.

Yet \textbf{training stability remains a critical and underexplored challenge}. In practice, convergence is sensitive to initialization; codebooks may stagnate in low-utilization phases; and abrupt utilization collapse can occur mid-training, particularly at scale. We argue that these failure modes are not incidental but symptomatic of a deeper structural issue: the Encoder--Decoder and the Codebook are entangled in their training, such that neither can reliably fulfill its own responsibility in isolation. This entanglement masks the latent dysfunction of each module, leaving underlying issues unresolved and making the overall system contingent on fragile inter-module cooperation rather than principled individual competence. In Section~\ref{sec:failure_modes}, we provide a principled analysis of the latent problems in existing VQ tokenizer training that this entanglement conceals.

We propose \textbf{StableVQ}, a set of lightweight interventions that enables principled, stable VQ training without relying on fragile inter-module cooperation. Our contributions are as follows:

\begin{itemize}

    \item \textbf{A separation-of-concerns analysis of VQ tokenizer training.} We revisit the proper responsibility of each module in VQ training and show that inter-module entanglement has long concealed latent dysfunctions in each. This analysis reframes training instability as a failure of modular responsibility rather than a fundamental limitation of the quantization paradigm.

    \item \textbf{StableVQ: principled and lightweight interventions for stable training.} Guided by the above analysis, we propose three targeted, parameter-free components that enable each module to fulfill its own responsibility independently. Together, they achieve principled, stable VQ training across diverse codebook sizes and initialization settings, without relying on heuristic design choices.

    \item \textbf{A more accessible performance ceiling for VQ tokenizers.} By eliminating the dependence on heuristic initialization and shared-projection architecture design, StableVQ allows the full expressive potential of VQ tokenizers to be realized without optimization artifacts standing in the way. A single linear projection suffices to reach state-of-the-art quality across diverse settings, and the principled modular stability established by our framework offers a solid theoretical footing for extending VQ training reliably to more demanding scenarios.

\end{itemize}
\section{Related Work}
\label{sec:related}

We provide a brief overview of related work here, with a more comprehensive version in Appendix~\ref{sup:related}. Vector-quantized representation learning was introduced by VQ-VAE~\cite{vqvae}, which maps continuous encoder features to discrete code indices through nearest-neighbor lookup in a learned codebook. Subsequent tokenizers improve reconstruction quality and token capacity through hierarchical latents~\cite{vqvae2}, perceptual and adversarial objectives~\cite{vqgan}, residual or multi-stage quantization~\cite{rqvae}, and ViT-based architectures~\cite{vitvqgan}. A central challenge in these systems is codebook collapse and low utilization, especially as codebook size or embedding dimension increases. Existing remedies include code reset or replacement~\cite{cvq}, low-dimensional embeddings and normalization~\cite{vitvqgan}, relaxed or soft-assignment paths~\cite{dalle,gumbel-softmax,ibq}, and scalar or binary quantization alternatives~\cite{fsq,lfq}. More recent shared-projection methods such as VQ-STE++, SimVQ, and FVQ~\cite{straighten,simvq,fvq} reparameterize code vectors through a shared function, substantially improving utilization with little architectural overhead. StableVQ builds on this line and studies the remaining training-instability problem.

\section{Background}
\label{sec:background}

\subsection{Vector Quantization}

Given an encoder $E$ and a decoder $D$, a VQ-VAE~\cite{vqvae} maps an input image $\mathbf{x}$ to a spatial feature map $\mathbf{z} = E(\mathbf{x}) \in \mathbb{R}^{H \times W \times d}$. Each spatial feature $\mathbf{z}_{ij} \in \mathbb{R}^d$ is quantized by nearest-neighbor lookup in a codebook $\mathcal{C} = \{\mathbf{e}_k\}_{k=1}^{K}$:
\begin{equation}
    k^*_{ij} = \arg\min_{k \in [K]} \|\mathbf{z}_{ij} - \mathbf{e}_k\|_2^2, \qquad \hat{\mathbf{z}}_{ij} = \mathbf{e}_{k^*_{ij}}.
\end{equation}
The training objective decomposes into three terms:
\begin{equation}
    \mathcal{L} = \mathcal{L}_{\text{recon}} + \beta \underbrace{\|\mathbf{z} - \text{sg}[\hat{\mathbf{z}}]\|^2}_{\text{commitment loss}} + \underbrace{\|\text{sg}[\mathbf{z}] - \hat{\mathbf{z}}\|^2}_{\text{VQ loss}},
    \label{eq:vq_loss}
\end{equation}
where $\text{sg}[\cdot]$ is the stop-gradient operator. The commitment loss constrains encoder outputs to remain near their assigned codes; the VQ loss drives codebook entries toward the encoder output distribution; and the end-to-end gradient from the reconstruction loss is passed back to the Encoder via the Straight-Through Estimator (STE), which approximates $\partial \mathcal{L}/\partial \mathbf{z}_{ij} \approx \partial \mathcal{L}/\partial \hat{\mathbf{z}}_{ij}$.

\subsection{Codebook with a Shared Projection}
\label{sec:shared_distribution}

In the standard formulation, each code $\mathbf{e}_k$ is an independent parameter, so the VQ loss gradient is nonzero only for selected codes---a property we term \emph{gradient sparsity}. As codebook size $K$ grows, the fraction of codes updated per step diminishes, exacerbating collapse risk.

A recent line of work addresses gradient sparsity via a \emph{shared projection}: codes are reparameterized as $\tilde{\mathbf{e}}_k = f_\theta(\mathbf{e}_k)$, where $f_\theta$ is a differentiable function shared across all codes:
\begin{equation}
    k^*_{ij} = \arg\min_{k \in [K]} \|\mathbf{z}_{ij} - f_\theta(\mathbf{e}_k)\|_2^2.
    \label{eq:shared_proj}
\end{equation}
Because $f_\theta$ is shared, gradients from any selected code propagate to influence all $\{\mathbf{e}_k\}$, transforming VQ training into an alignment problem between the projected code distribution $\mathcal{P}_{\mathcal{C}} = \{f_\theta(\mathbf{e}_k)\}_{k=1}^K$ and the encoder output distribution $\mathcal{P}_{\mathcal{Z}} = \{E(\mathbf{z}_{ij})\}$. 

Representative instantiations of this paradigm include the affine reparameterization approach of~\cite{straighten}, which applies a shared learnable scale-and-shift transformation to the code vectors, rescaling each quantized embedding as $q = c_{\mathrm{mean}} + c_{\mathrm{std}} \odot \hat{q}$ where $c_{\mathrm{mean}}$ and $c_{\mathrm{std}}$ are codebook-wide affine parameters; SimVQ~\cite{simvq}, which defines $f_	\theta$ as a learnable linear layer $W$ acting on a fixed latent basis, so that each code vector is produced as $c_k = q_k W$, optimizing the entire linear space spanned by the codebook rather than individual code vectors; and FVQ~\cite{fvq}, which employs a more expressive nonlinear projector to remap code vectors, enabling full codebook utilization. These methods substantially improve codebook utilization, but training stability remains unsolved, as shown below.

\section{Method}
\label{sec:method}

\subsection{Failure Modes of VQ Training}
\label{sec:failure_modes}

\paragraph{Limitations of shared-projection methods.}
Although shared-projection methods substantially mitigate gradient sparsity, they do not fully resolve the challenge of codebook distribution alignment. First, gradient propagation through $f_\theta$ influences all codes indirectly: the signal reaching inactive codes is diffuse and undirected, providing no guarantee that they converge toward the correct regions of the token distribution. Second, this indirect influence is subject to decay: once a small subset of codes covers the token distribution well enough to minimize the VQ loss, the training signal driving the remaining codes becomes negligible. Furthermore, shared-projection methods operate purely on the codebook side and do not account for potential instabilities in the Encoder's optimization during training---a separate source of fragility that can compound with codebook misalignment and destabilize the system as a whole.

\begin{figure}[t]
    \centering
    \includegraphics[width=\linewidth]{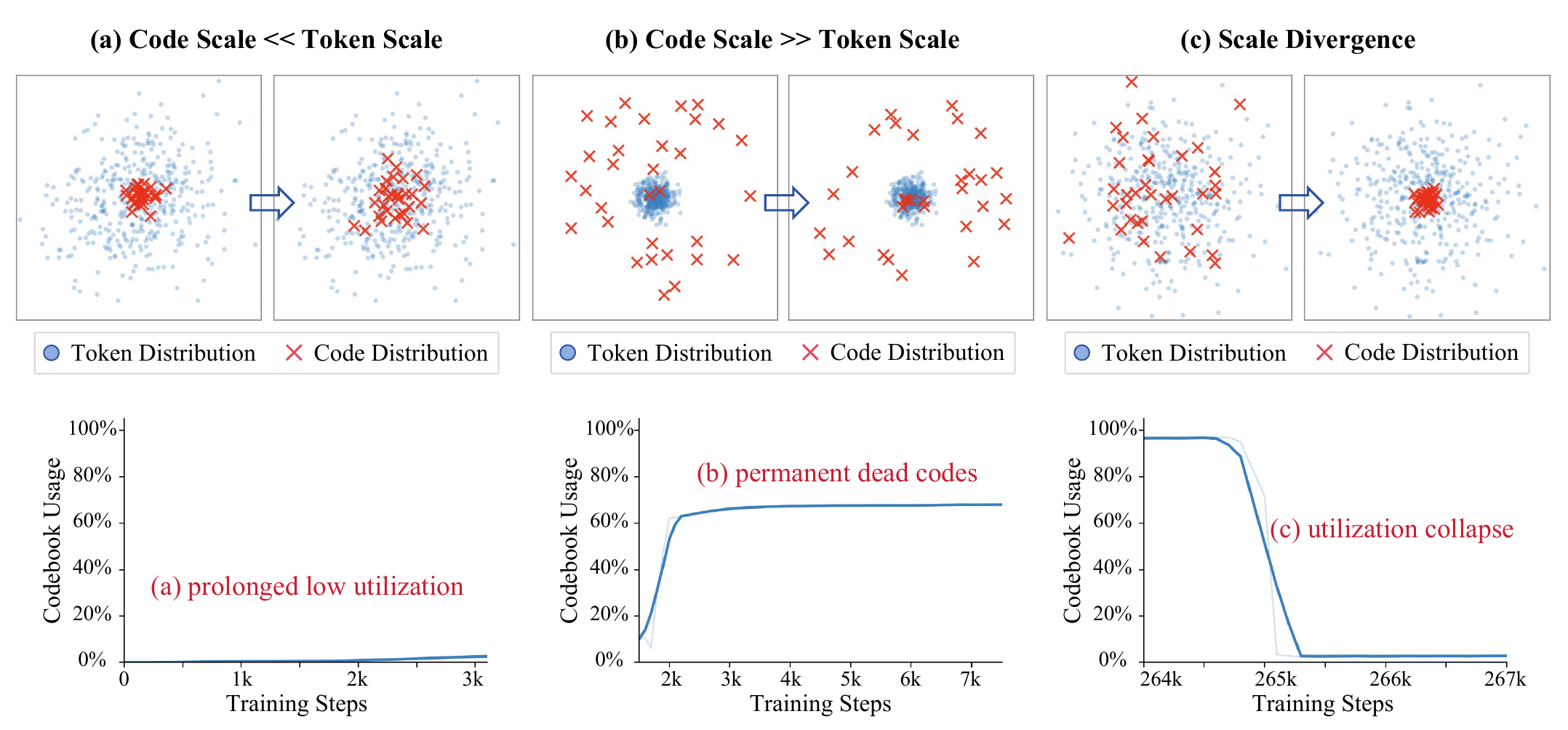}
    \vspace{-10pt}
    \caption{
        Three characteristic failure modes of VQ training, illustrated through the relationship between the token distribution (blue) and code distribution (red) at different training stages.
    }
    \label{fig:failure_modes}
\end{figure}

\paragraph{Challenging failure modes.}
Wasserstein VQ~\cite{wassersteinvq} analyzes static relationships between token and code distributions. We further examine characteristic failure modes observed during training, analyzing how the token--code distributional relationship evolves in each case and what drives this evolution (Figure~\ref{fig:failure_modes}). \textbf{(a) Code scale $\ll$ Token scale}: when the code distribution occupies a smaller scale region than the token distribution, the Codebook receives only sparse optimization targets due to low utilization, while the token distribution fluctuates unpredictably under the competing gradients of the STE-passed reconstruction signal and the commitment loss. The system thus falls into \emph{prolonged low utilization}; when these fluctuations are severe enough, commitment loss spikes can escalate to NaN gradients before the codebook ever reaches meaningful utilization. 
\textbf{(b) Code scale $\gg$ Token scale}: when the code distribution spans a much larger region than the token distribution, utilization rises rapidly as codes within the token scale region are quickly activated. However, as more in-range codes are claimed, the commitment loss and VQ loss signals become increasingly saturated, rapidly diminishing the training signal for out-of-range codes. This leaves the majority of the codebook virtually unreachable by nearest-neighbor assignment, resulting in \emph{permanent dead codes}. 
\textbf{(c) Scale divergence}: even in a well-utilized codebook, the two distributions are in continuous dynamic alignment as the reconstruction loss drives the token distribution to evolve. Should the Codebook momentarily fail to track a sudden distributional shift, the erroneous STE gradients passed to the Encoder tend to amplify the divergence rather than correct it, triggering a positive-feedback loop that rapidly escalates the scale mismatch and causes \emph{utilization collapse} instantaneously.

The failure modes described above share a common root cause: the Encoder and Codebook are not given the conditions to independently fulfill their own responsibilities. Guided by the principle of separation of concerns, we analyze the proper learning objective of each module and identify where the current training pipeline prevents each from fulfilling its own role.

\subsection{Dynamic Straight-Through Estimator}
\label{sec:dynamic_ste}

\paragraph{Gradient Estimation Gap.}
The Encoder's proper learning objective is to optimize the reconstruction space under the discrete code constraint imposed by the commitment loss. Ideally, the commitment loss and the STE-passed reconstruction gradient should cooperate toward this objective.
However, when tokens are assigned to distant codes, the STE gradient becomes an unreliable estimate of the true reconstruction direction. Such unreliable gradients can conflict with the commitment loss, push the token and code distributions further apart, and trigger commitment-loss spikes that may escalate to NaN values. By amplifying distributional errors rather than correcting them, they also become a primary driver of sudden utilization collapse, undermining overall training robustness.

\paragraph{Gradient Quality Weighting.}
The key observation is that when multiple tokens hit the same code, the \emph{relatively farther} tokens are those whose STE gradients are most unreliable. We therefore assign each token a gradient weight based on its distance to the assigned code relative to the best-matched token for that code:
\begin{equation}
    w_{ij} = \mathrm{sg}\!\left[\frac{d^*_{k^*_{ij}}}{\|\mathbf{z}_{ij} - f_\theta(\mathbf{e}_{k^*_{ij}})\|_2^2}\right] \in (0, 1], \quad d^*_{k} = \min_{(i',j')} \|\mathbf{z}_{i'j'} - f_\theta(\mathbf{e}_k)\|_2^2,
    \label{eq:rqr}
\end{equation}
where $d^*_k$ is the minimum squared distance from code $k$ to any token in the current batch. When a token is the best-matched token for its assigned code, it receives the full gradient with $w_{ij} = 1$. As its relative quantization error grows, the weight decreases accordingly.

We modulate the STE by this per-token weight:
\begin{equation}
    \hat{\mathbf{z}}_{ij} = w_{ij} \cdot \mathbf{z}_{ij} + \text{sg}[\hat{\mathbf{z}}_{ij} - w_{ij} \cdot \mathbf{z}_{ij}].
    \label{eq:dynamic_ste}
\end{equation}
This formulation attenuates the STE gradient for relatively distant tokens while preserving the full gradient for the token with the best assignment. Crucially, when all tokens in a batch are assigned to well-matched codes (high utilization, stable training), $w_{ij} \approx 1$ for all tokens and Dynamic STE reduces to the standard STE. The intervention is thus self-deactivating under healthy training conditions, and no threshold hyperparameter is required.

\begin{figure}[t]
    \centering
    \includegraphics[width=\linewidth]{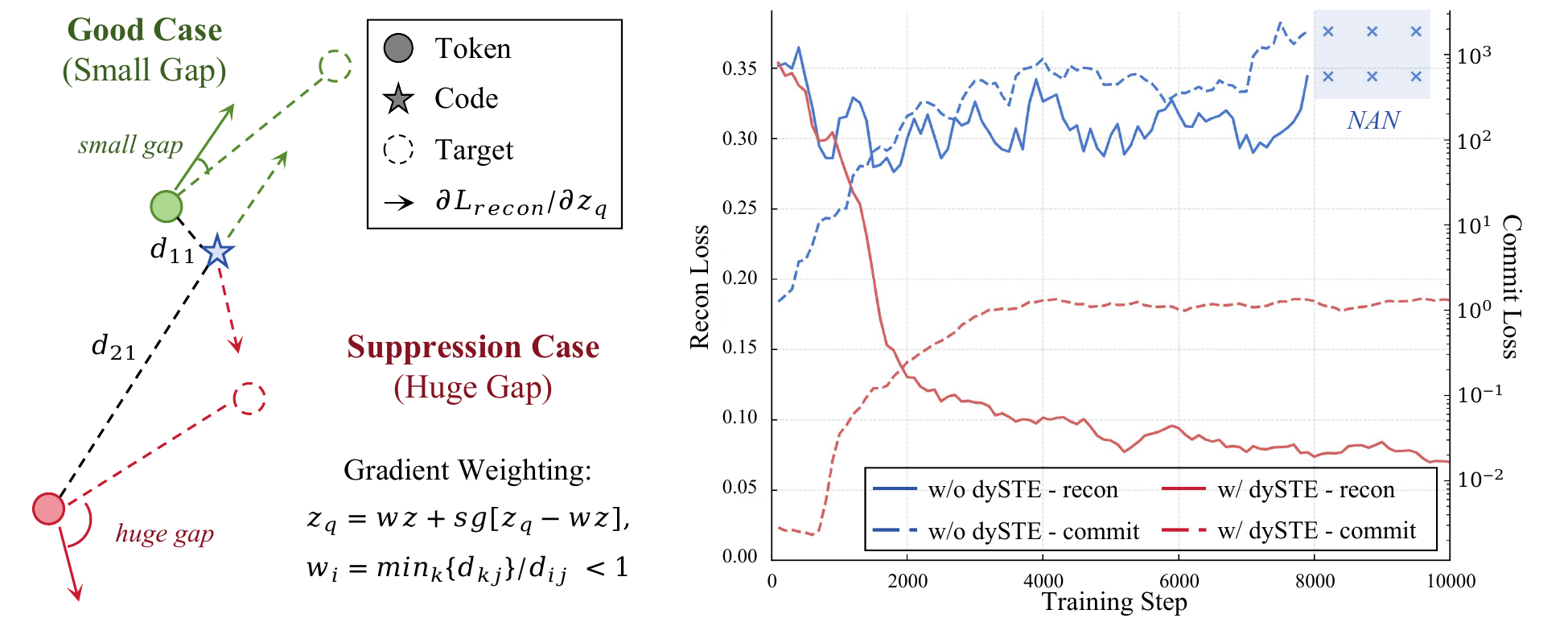}
    \vspace{-10pt}
    \caption{
        \textbf{Left: Dynamic STE design.} Relatively farther tokens have more unreliable gradients and receive proportionally suppressed contributions.
        \textbf{Right: Pilot Study.} Standard STE causes commitment loss spikes and training collapse, whereas Dynamic STE maintains stable losses.
    }
    \vspace{-5pt}
    \label{fig:pilot_ste}
\end{figure}

\paragraph{Pilot Study.}
To verify the effect of Dynamic STE on the Encoder's learning objective, we freeze the Codebook and train only the Encoder--Decoder. Figure~\ref{fig:pilot_ste} shows that standard STE leads to severe commitment-loss spikes and reconstruction-loss oscillations, ultimately causing losses to diverge to NaN, while Dynamic STE suppresses these instabilities and keeps losses stable throughout training. This confirms that the gradient estimation gap introduces latent instability into Encoder training.

\subsection{Region VQ Loss}
\label{sec:region_vq}

\paragraph{Unguaranteed Distribution Alignment.}
The Codebook's proper learning objective is to track the encoder output distribution through a clean and independent optimization process. In the ideal setting, this should endow the Codebook with the ability to guarantee codebook utilization on its own, without relying on assistance from the Encoder. However, the standard VQ loss only provides direct learning targets to the codes selected in the current step, leaving the majority of codes without explicit supervision. Although shared-projection methods allow gradients to reach all codes through $f_\theta$, the resulting signal remains indirect and undirected, and is inherently subject to attenuation during training. Consequently, the Codebook cannot independently guarantee full utilization; instead, activation of the remaining codes often depends on unstable fluctuations in the encoder output, meaning that codebook utilization is not reliably guaranteed in practice.

\paragraph{Asymmetry in VQ Loss.}
The standard VQ loss is inherently asymmetric: every token receives an explicit target through the commitment loss, whereas only selected codes are assigned meaningful objectives. Letting all codes take the nearest token as their target seems a natural remedy, yet this often fails to provide correct learning directions. The solution lies in shifting from point-wise to distribution-wise alignment. Since asymmetry arises from many tokens selecting few codes, we propagate the targets received by active codes to nearby inactive ones, which we term Region VQ.

\begin{figure}[t]
    \centering
    \includegraphics[width=\linewidth]{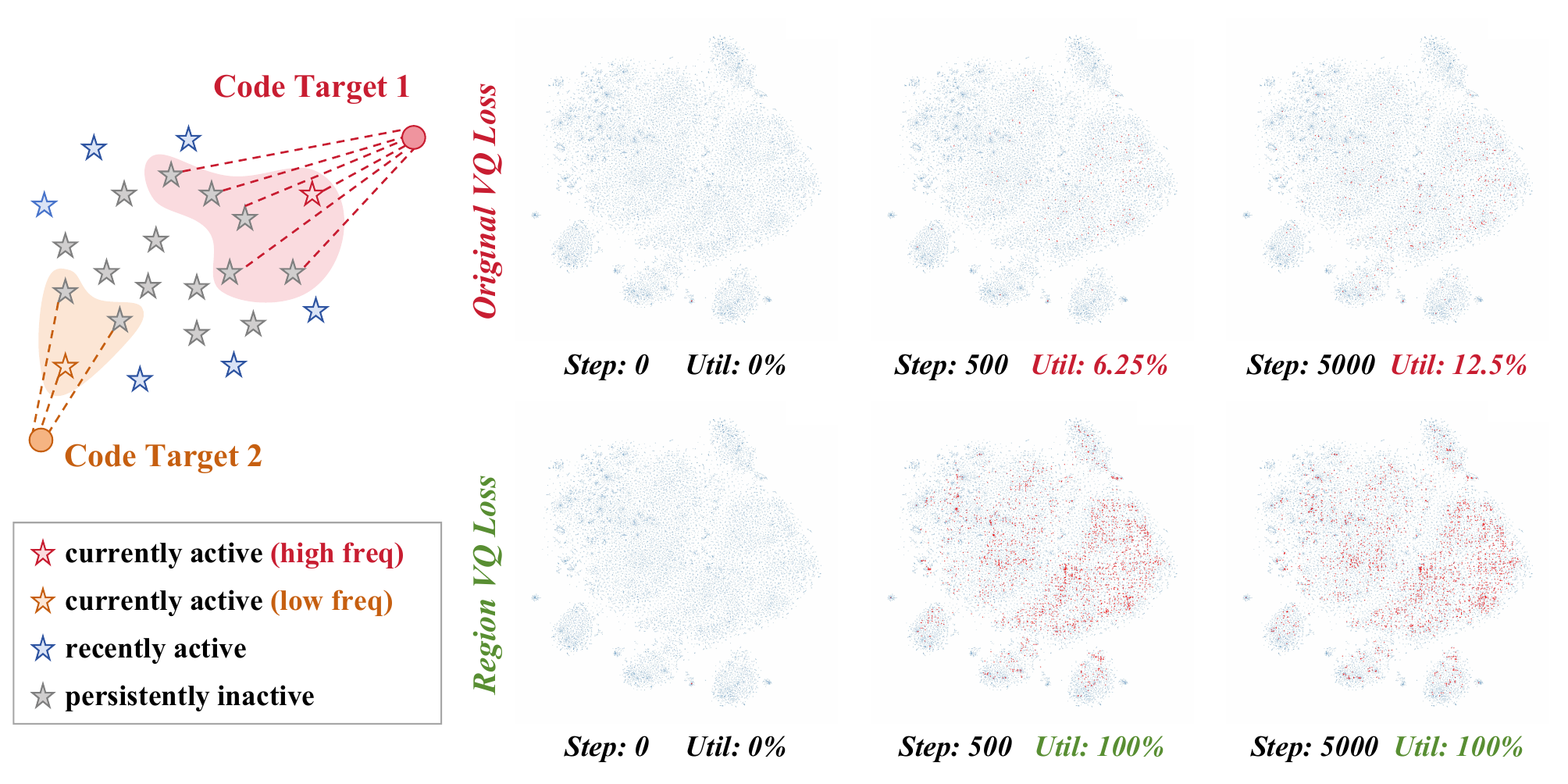}
    \vspace{-10pt}
    \caption{
        \textbf{Left: Region VQ Loss.} Active codes propagate targets to nearby inactive codes proportionally to selection count.
        \textbf{Right: Pilot Study.} T-SNE visualizations at different training steps (blue: encoder outputs; red: codebook entries) with codebook utilization.
    }
    \label{fig:pilot_region}
\end{figure}

\paragraph{Code Target Assignment.}
Let $[K]=\{1,\ldots,K\}$ denote all code indices and $\mathcal{S}_t\subseteq[K]$ denote the set of codes selected at step $t$. To distinguish codes by activity, we maintain a FIFO queue of length $W$, which defines the window-active set $\mathcal{A}_t = \bigcup_{s=t-W+1}^{t} \mathcal{S}_s$ and the persistently inactive set $\mathcal{N}_t = [K] \setminus \mathcal{A}_t$. Let $\mathcal{U}_k$ contain the token features $\mathbf{z}_u$ assigned to code $k$, with $n_k=|\mathcal{U}_k|$. Each source code $k\in\mathcal{S}_t$ receives a quota $q_k\propto n_k$ and propagates its target to the $q_k$ nearest codes in $\mathcal{N}_t$, denoted $\mathcal{R}_k$ (Algorithm~\ref{alg:region_vq}). Let $\mathcal{K}_j = \{k\in\mathcal{S}_t : j \in \mathcal{R}_k\}$ denote the sources propagating to code $j$. Recently active codes $\mathcal{A}_t\setminus\mathcal{S}_t$ and unclaimed inactive codes $\{j\in\mathcal{N}_t:|\mathcal{K}_j|=0\}$ retain self-targets and yield zero loss, while other effective code targets are defined as follows:
\begin{equation}
    \mathbf{t}_j =
    \begin{cases}
        \displaystyle \frac{1}{|\mathcal{U}_j|} \sum_{u \in \mathcal{U}_j} \mathbf{z}_u,
        & j \in \mathcal{S}_t, \\[4pt]
        \displaystyle \frac{1}{|\mathcal{K}_j|} \sum_{k \in \mathcal{K}_j} \mathbf{t}_k,
        & j \in \mathcal{N}_t,\; |\mathcal{K}_j| > 0.
    \end{cases}
    \label{eq:code_targets}
\end{equation}

Let $\mathcal{M}_t = \mathcal{S}_t \cup \{j \in \mathcal{N}_t : |\mathcal{K}_j| > 0\}$ denote codes with effective targets. The unified codebook update loss is
\begin{equation}
    \mathcal{L}_{\text{code}} = \frac{1}{|\mathcal{M}_t|} \sum_{k \in \mathcal{M}_t} \|f_\theta(\mathbf{e}_k) - \text{sg}[\mathbf{t}_k]\|_2^2.
    \label{eq:region_vq_loss}
\end{equation}

\paragraph{Pilot Study.}
To isolate the effect of Region VQ Loss on the Codebook's learning objective, we freeze the Encoder and optimize only the Codebook, which uses a two-layer ViT Block as the shared projector to ensure sufficient learning capacity. Figure~\ref{fig:pilot_region} shows T-SNE visualizations of the encoder output distribution and codebook entries at Steps 0, 500, and 5000. The standard VQ loss stagnates at around 12.5\% utilization even after 5000 steps, whereas Region VQ Loss reaches full utilization by Step 500 and maintains it throughout training. This confirms that the unguaranteed distribution alignment problem is intrinsic to the VQ loss objective, and that Region VQ Loss directly resolves it by providing every code with a principled learning target.

\subsection{Decoupled Schedule}
\label{sec:decoupled_schedule}

\paragraph{Coupled Optimization.}
The Encoder--Decoder and the Codebook have fundamentally different optimization characteristics and should be governed by independent learning rate schedules. The Encoder benefits from warmup-plus-annealing to stabilize its complex multi-objective landscape. The Codebook, whose task is to continuously track the evolving encoder distribution, requires sustained high learning rates especially during the early phase when the encoder output distribution changes most rapidly. Coupling both under either a constant learning rate or a warmup-plus-annealing schedule leads to suboptimal performance or reduced codebook utilization. 

\paragraph{Objective-Driven Schedule Decoupling.}
Prior works~\cite{straighten,fvq} have shown that a warmup-plus-annealing learning rate schedule benefits VQ training quality, yet it often leads to degraded codebook utilization. To compensate, FVQ introduces a more expressive shared projector to maintain utilization under this schedule. Our preceding analysis reveals that this tension stems from a more fundamental issue: the Encoder--Decoder and the Codebook have inherently different learning objectives, and therefore require distinct optimization schedules. We treat them as two independent optimization systems, each scheduled according to its own objective:

\begin{itemize}
    \item \textbf{Encoder--Decoder} is responsible for reconstruction under discrete regularization, a complex multi-objective task that benefits from warmup-plus-annealing to stabilize early optimization and ensure smooth convergence.
    \item \textbf{Codebook} is responsible for continuously tracking the encoder output distribution, a clean and dedicated objective for which a constant high learning rate with no warmup may be most beneficial, ensuring adequate gradient magnitude from the very first step.
\end{itemize}

\paragraph{Pilot Study.}

\begin{figure}[t]
    \centering
    \includegraphics[width=\linewidth]{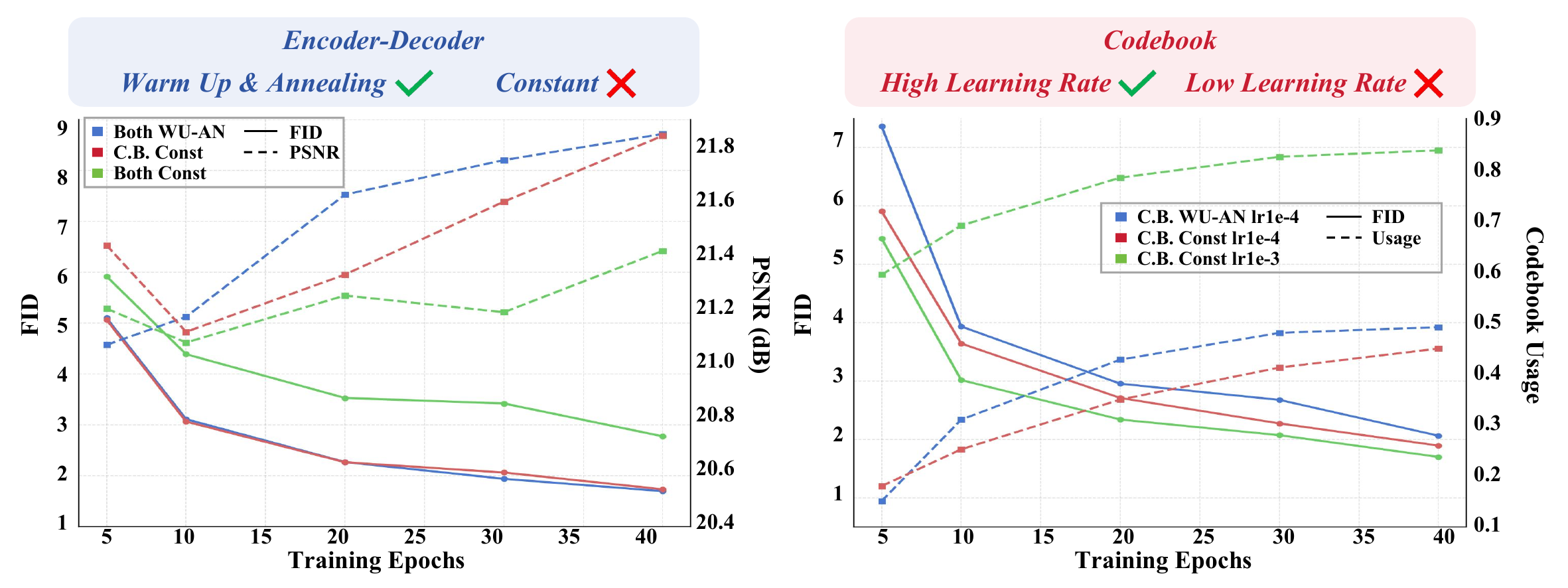}
    \vspace{-10pt}
    \caption{
        Pilot study on learning rate schedules for different modules. WU-AN denotes Warmup-plus-Annealing; C.B. denotes Codebook.
    }
    \label{fig:pilot_schedule}
    \vspace{-10pt}
\end{figure}

We conduct two controlled experiments to verify this design (Figure~\ref{fig:pilot_schedule}). Using the FVQ architecture (left), we ablate which module benefits from warmup-plus-annealing: setting the Codebook to a constant learning rate (red) incurs no performance degradation, whereas applying a constant rate to the Encoder--Decoder leads to a clear quality drop. Using the SimVQ architecture (right), we examine what schedule the Codebook requires: utilization is not improved by complex schedules, but benefits from a stable and sufficiently high constant learning rate.

\section{Experiments}
\label{sec:experiments}

\subsection{Main Results}
\label{sec:main_results}

\paragraph{Setup.}
We evaluate StableVQ on ImageNet~\cite{imagenet} at $256 \times 256$ resolution using a VQGAN-style~\cite{vqgan} encoder--decoder with downsampling factor $f=16$, producing $16\times16=256$ tokens per image. We evaluate reconstruction quality by rFID and LPIPS, and codebook utilization as the fraction of activated codes, on the ImageNet validation set.
We compare against a range of baselines, with particular focus on shared-projection methods SimVQ~\cite{simvq} and FVQ~\cite{fvq}. For these baselines, we re-implement their results following their respective original configurations, and evaluate using on-the-fly reconstruction rather than a save-then-reload pipeline to ensure fair and accurate metric computation. StableVQ uses the simplest single-layer linear shared projector by default.

\paragraph{Reconstruction results.}

\begin{table}[t]
\caption{Reconstruction results on ImageNet $256\times256$ with $16\times16$ tokens.
$^\dagger$ denotes a larger encoder--decoder. UR-AUC denotes Usage Recovery AUC under the robustness test.
}
\label{tab:main_recon}
\centering
\small
\setlength{\tabcolsep}{3.5pt}
\renewcommand{\arraystretch}{1.15}
\begin{tabular}{cccccccc}
\toprule
Method & Projector & Epochs & Codebook Size ($n \times d$) & rFID$\downarrow$ & LPIPS$\downarrow$ & Usage$\uparrow$ & UR-AUC$\uparrow$ \\
\midrule

LlamaGen~\cite{llamagen} & --- & 40 & $16{,}384 \times 8$    & 2.19 & 0.2281 & 97\%   & \multirow{2}{*}{---} \\
LlamaGen~\cite{llamagen} & --- & 40 & $16{,}384 \times 256$  & 9.21 & ---    & 0.29\% &  \\

IBQ$^\dagger$~\cite{ibq}  & --- & 330 & $16{,}384 \times 256$  & 1.37 & 0.2235 & 96\%   & \multirow{2}{*}{---} \\
IBQ$^\dagger$~\cite{ibq}  & --- & 330 & $262{,}144 \times 256$ & 1.00 & 0.2030 & 84\%   &  \\

\midrule
\multicolumn{8}{l}{\textit{Shared-Projection-Based Methods}} \\
\midrule

VQGAN-LC~\cite{vqgan-lc} & Linear-1   & 20 & $16{,}384 \times 8$    & 3.01 & 0.2358 & 99\%  & \multirow{2}{*}{---} \\
VQGAN-LC~\cite{vqgan-lc} & Linear-1   & 20 & $100{,}000 \times 8$   & 2.62 & 0.2212 & 99\%  &  \\

SimVQ~\cite{simvq}    & Linear-1   & 40 & $16{,}384 \times 256$  & 2.89 & 0.2492 & 100\% & \multirow{2}{*}{$2.17{\scriptstyle\pm0.32}$} \\
SimVQ~\cite{simvq}    & Linear-1   & 40 & $262{,}144 \times 256$ & 3.16 & 0.2516 & 100\% &  \\

FVQ~\cite{fvq}      & ViTBlock-2 & 40 & $16{,}384 \times 256$  & 1.70 & 0.2176 & 100\% & \multirow{2}{*}{$8.08{\scriptstyle\pm0.24}$} \\
FVQ~\cite{fvq}      & ViTBlock-2 & 40 & $262{,}144 \times 256$ & 1.29 & 0.2003 & 100\% &  \\

\midrule

StableVQ & Linear-1 & 40 & $16{,}384 \times 256$  & 1.22 & 0.2235 & 100\% & \multirow{4}{*}{$\mathbf{60.59}{\scriptstyle\pm1.52}$} \\
StableVQ & Linear-1 & 40 & $262{,}144 \times 256$  & 1.05 & 0.1947 & 100\% &  \\

StableVQ & Linear-1 & 120 & $16{,}384 \times 256$  & \textbf{1.13} & \textbf{0.2134} & 100\% &  \\
StableVQ & Linear-1 & 120 & $262{,}144 \times 256$  & \textbf{0.92} & \textbf{0.1893} & 100\% &  \\

\bottomrule
\end{tabular}
\vspace{-5pt}
\end{table}

Table~\ref{tab:main_recon} presents reconstruction results.
While SimVQ and FVQ both achieve 100\% utilization under their respective standard configurations, each comes with notable limitations. SimVQ adopts a constant learning rate to sustain full utilization, but the lack of annealing results in substantially degraded reconstruction quality. FVQ relies on a carefully engineered projector architecture---including ViT block depth and patch size---to maintain utilization under warmup-plus-annealing, and the patch embedding operation constrains the codebook size to perfect squares.

StableVQ achieves superior reconstruction quality with a single linear projection layer, matching or surpassing methods that rely on more complex projectors. It requires no projector-specific design and imposes no structural constraints, while maintaining full utilization across a broader range of challenging scenarios as the ablation studies demonstrate.
When VQ training stability is no longer the bottleneck, the optimization strategy becomes the dominant factor in reconstruction quality;
StableVQ uses a discriminator following prior works~\cite{stylegan-t,rae} as its adversarial supervision.

\paragraph{Robustness test.}

\begin{figure}[t]
    \centering
    \includegraphics[width=\linewidth]{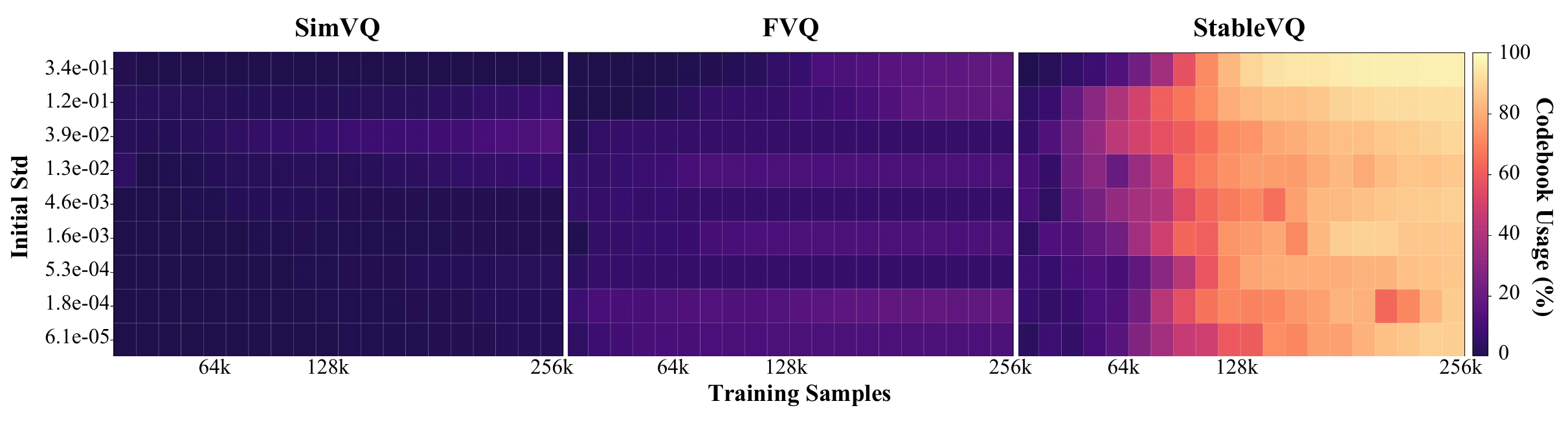}
    \vspace{-20pt}
    \caption{Codebook usage recovery under different codebook-token distribution relationships. 
    }
    \vspace{-10pt}
    \label{fig:usage_recovery}
\end{figure}

To further test the stability of different methods, we introduce UR-AUC in Table~\ref{tab:main_recon}, which denotes Usage Recovery AUC. It measures codebook utilization recovery under codebook-token distribution mismatch and is computed as the average AUC of codebook usage curves over the tested mismatch settings (Appendix~\ref{sup:robustness_test}). As visualized in Figure~\ref{fig:usage_recovery}, SimVQ and FVQ recover usage slowly and only under limited mismatch conditions, whereas StableVQ rapidly restores full codebook usage across diverse codebook-token distribution relationships. This indicates that StableVQ avoids prolonged low utilization and dead-code issues under different training conditions, providing strong robustness guarantees for scaling VQ training and applying it to broader scenarios.

\paragraph{Generation results.}

Following the IBQ~\cite{ibq} generation setup, we train class-conditional autoregressive transformers on StableVQ tokens. 
Table~\ref{tab:main_gen} shows competitive ImageNet $256\times256$ generation results, with complete baseline comparisons provided in Appendix~\ref{sup:results}.

\begin{table}[h]
\caption{Class-conditional image generation on ImageNet $256\times256$.}
\label{tab:main_gen}
\centering
\small
\setlength{\tabcolsep}{8pt}
\begin{tabular}{ccccccccc}
    \toprule
    Type & Tokenizer & Generator & Param. & FID$\downarrow$ & IS$\uparrow$ & Pre.$\uparrow$ & Rec.$\uparrow$ \\
    \midrule
    Vanilla AR     & IBQ~\cite{ibq}      & IBQ-B~\cite{ibq}         & 342M & 2.88  & 254.7 & \textbf{0.84} & 0.51 \\
    Vanilla AR     & IBQ~\cite{ibq}      & IBQ-L~\cite{ibq}         & 649M & 2.45  & \textbf{267.5} & \textbf{0.83} & 0.52 \\
    \midrule
    Vanilla AR     & StableVQ & IBQ-B~\cite{ibq}   & 342M & \textbf{2.35} & \textbf{256.0} & 0.82 & \textbf{0.58} \\
    Vanilla AR     & StableVQ & IBQ-L~\cite{ibq}   & 649M & \textbf{2.18}  & 250.4   & 0.82  & \textbf{0.59}  \\
    \bottomrule
\end{tabular}
\vspace{-5pt}
\end{table}

\subsection{Ablation Studies}
\label{sec:ablation}

The main results above are obtained under each method's standard setting, where methods such as SimVQ and FVQ can also reach full utilization. However, these methods still do not resolve the fundamental deficiencies analyzed in Section~\ref{sec:method}. We therefore turn to more challenging yet common settings to evaluate training robustness.

\paragraph{Ablation 1: Codebook expansion.}

We first set the experiment to the case of Figure~\ref{fig:failure_modes}(a), where the Codebook is initialized within a very small numerical range---the most common initialization trick in VQ training. In this setting, we use a single linear layer as the shared projector and adopt a warmup-plus-annealing schedule with peak learning rate $1\mathrm{e}{-4}$. We report peak commitment loss, codebook utilization, and reconstruction metrics, and evaluate different combinations of the strategies proposed in Section~\ref{sec:method} on top of the baseline for a more complete analysis. When Decoupled Schedule is used, the Codebook learning rate is set to a constant $1\mathrm{e}{-3}$.

\begin{table}[t]
\caption{Ablation on proposed strategies under the codebook expansion setting.}
\label{tab:ablation1}
\centering
\small
\setlength{\tabcolsep}{6pt}
\begin{tabular}{ccc cccccc}
\toprule
Region VQ & Dyn. STE & De. Sch. & Peak Commit & Util.$\uparrow$ & rFID$\downarrow$ & LPIPS$\downarrow$ & PSNR$\uparrow$ & SSIM$\uparrow$ \\
\midrule
           &            &            & 53.11   & 49.13\% & 2.06 & 0.2297 & 21.65 & 0.5730 \\
\checkmark &            &            & $>$200 (NaN)  & ---   & ---  & ---    & ---   & ---    \\
           & \checkmark &            & $<$0.1  & 1.27\%  & 6.68 & 0.3100 & 19.89 & 0.5026 \\
           &            & \checkmark & $<$0.1  & 83.77\% & 1.71 & 0.2209 & 21.79 & 0.5864 \\
           & \checkmark & \checkmark & $<$0.1  & 21.66\% & 2.07 & 0.2354 & 21.36 & 0.5665 \\
\checkmark & \checkmark &            & $<$0.1  & \textbf{100\%}   & 1.72 & \textbf{0.2200} & 21.84 & 0.5854 \\
\checkmark &            & \checkmark & $<$0.1  & \textbf{100\%}   & 1.75 & 0.2206 & \textbf{21.85} & 0.5860 \\
\checkmark & \checkmark & \checkmark & $<$0.1  & \textbf{100\%}   & \textbf{1.70} & 0.2208 & 21.81 & \textbf{0.5879} \\
\bottomrule
\end{tabular}
\vspace{-5pt}
\end{table}

The results are consistent with the analysis in Section~\ref{sec:method}. Region VQ Loss alone causes NaN collapse because the Codebook still updates too slowly under the shared warmup schedule. Dynamic STE suppresses commitment-loss spikes and stabilizes the Encoder, but utilization remains very low, revealing that conventional training relies on unstable encoder oscillations to activate codes. Decoupled Schedule improves distribution tracking from the start, yet still falls short of full utilization under the standard VQ loss. Once Region VQ Loss is combined with either Dynamic STE or Decoupled Schedule, full utilization is recovered with strong reconstruction quality. Using all three components gives the most complete solution, jointly ensuring Encoder stability, dense Codebook supervision, and fast distribution tracking.

\paragraph{Ablation 2: Codebook shrinkage.}

\begin{table}[t]
\caption{Ablation on methods under the codebook shrinkage setting.}
\label{tab:ablation2}
\centering
\small
\setlength{\tabcolsep}{8pt}
\begin{tabular}{cccccccc}
    \toprule
    Projector & Region VQ & Init & Util.$\uparrow$ & rFID$\downarrow$ & PSNR$\uparrow$ & SSIM$\uparrow$ & LPIPS$\downarrow$ \\
    \midrule
    ViTBlock-2~\cite{fvq} &            & uniform  & 18.75\% & 2.4355 & 20.9789 & 0.5677 & 0.2414 \\
    ViTBlock-2~\cite{fvq} &            & gaussian & 62.5\%  & 1.9530 & 21.3774 & 0.5836 & 0.2273 \\
    \midrule
    ViTBlock-2~\cite{fvq} & \checkmark & uniform  & \textbf{100\%}   & \textbf{1.8176} & \textbf{21.5497} & 0.5896 & \textbf{0.2210} \\
    ViTBlock-2~\cite{fvq} & \checkmark & gaussian & \textbf{100\%}   & 1.8966 & 21.4603 & \textbf{0.5931} & 0.2255 \\
    \bottomrule
\end{tabular}
\vspace{-5pt}
\end{table}

We further set the experiment to the case of Figure~\ref{fig:failure_modes}(b), where the Codebook occupies a large range. In VQ training, this commonly arises when the space is constrained by $\ell_2$ normalization: after normalization, either uniform or gaussian initialization distributes codes broadly across the space, whereas encoder outputs, reflecting the statistics of natural images, concentrate in a much smaller region. This naturally creates a codebook shrinkage scenario. In this experiment, we set the codebook dimension to 1024 and use a stronger ViTBlock shared projector, following the FVQ configuration.
The results show that FVQ fails to achieve full utilization under either codebook initialization, and its utilization is strongly affected by initialization. Adding Region VQ consistently reaches full utilization and improves reconstruction quality regardless of initialization. This further supports that StableVQ provides a more principled solution, enabling ideal codebook usage across different VQ training conditions.

\section{Conclusion}
\label{sec:conclusion}

StableVQ shows that the long-standing instability of VQ training is not a limitation of vector quantization itself, but a consequence of entangled optimization objectives. By restoring \emph{separation of concerns} between the Encoder--Decoder and the Codebook, StableVQ turns full codebook utilization from a fragile heuristic outcome into a principled property that can be directly guaranteed. We believe this provides an important foundation for extending VQ to broader scenarios and more challenging applications, where robust training is essential.

\section*{Acknowledgments}
This work was partially supported by the National Natural Science Foundation of China under Grant U25B2067.

\bibliographystyle{plainnat}
\bibliography{references}

\newpage

\appendix

\section{Related Work}
\label{sup:related}

\paragraph{Vector Quantization.}
Vector-quantized representation learning is introduced by VQ-VAE~\cite{vqvae}, which maps continuous encoder features to discrete code indices through nearest-neighbor lookup in a learned codebook. VQ-VAE-2~\cite{vqvae2} improves this framework with hierarchical latent maps, and VQGAN~\cite{vqgan} combines vector quantization with perceptual and adversarial objectives, making discrete visual tokenizers a standard interface for high-fidelity image synthesis. A large body of work improves reconstruction quality and token capacity by modifying the quantization structure, including residual or multi-stage quantization methods such as RQ-VAE~\cite{rqvae} and ViT-based tokenizer architectures such as ViT-VQGAN~\cite{vitvqgan}. Another line addresses codebook collapse and low utilization, which become increasingly severe when scaling codebook size or embedding dimension; representative solutions include codebook reset and replacement strategies~\cite{cvq}, low-dimensional code embeddings and normalization~\cite{vitvqgan}, and soft-assignment-based training strategies such as stochastic or Gumbel-softmax quantization~\cite{dalle,gumbel-softmax} and IBQ~\cite{ibq}, which use relaxed or soft-to-hard categorical paths to improve codebook gradients beyond the selected hard code. Scalar-quantization-based methods, including FSQ~\cite{fsq} and LFQ~\cite{lfq}, replace learned vector-codebook lookup with quantization over scalar or binary values, simplifying optimization and improving usage at scale while introducing different capacity trade-offs. Shared-projection methods such as VQ-STE++~\cite{straighten}, SimVQ~\cite{simvq}, and FVQ~\cite{fvq} reparameterize code vectors through a shared function, providing an elegant and lightweight way to address the long-standing low-utilization problem in VQ training. Recently, pretrained vision foundation models have also been used to improve visual tokenizers, for example by initializing, regularizing, or aligning tokenizer representations with vision-foundation-model features~\cite{vqgan-lc,vqrae}, which can strengthen semantic representation quality but is complementary to the training-stability problem studied in this work. StableVQ builds on the simple shared-projection foundation, but revisits the optimization responsibilities of the Encoder--Decoder and Codebook and targets the latent instability that remains even when codebook utilization has been substantially improved.

\paragraph{Image Generation.}
Image generation has been developed along both continuous and tokenized modeling paradigms. Early autoregressive models such as PixelCNN~\cite{pixelcnn} and iGPT~\cite{igpt} model images directly in pixel space, but their sequential generation cost and weak compression make high-resolution synthesis difficult. Discrete tokenizers alleviate this bottleneck by converting images into compact latent token sequences: VQGAN~\cite{vqgan} applies transformer-based autoregressive modeling in the VQ latent space, while VQ-VAE2~\cite{vqvae2}, RQ-Transformer~\cite{rqvae}, and related residual or hierarchical token models further explore multi-level discrete representations. Non-autoregressive and masked-token generators such as MaskGIT~\cite{maskgit}, MAGE~\cite{mage}, and MAGVIT-v2~\cite{lfq} predict missing visual tokens and refine them iteratively, improving sampling efficiency and demonstrating the importance of tokenizer quality for generation. More recently, large-scale autoregressive image generators such as LlamaGen~\cite{llamagen}, VAR~\cite{var}, RandAR~\cite{randar}, and Open-MAGVIT2~\cite{openmagvit2} show that language-model-style next-token, next-scale, or randomized-order prediction can achieve strong visual synthesis when paired with expressive visual tokens. In parallel, diffusion, score-based, and flow-matching models~\cite{ddpm,score_sde,flow_matching,latent_diffusion,dit} generate images through continuous denoising or transport processes, and hybrid alternatives such as MAR~\cite{mar} reduce or remove the dependence on hard vector quantization. These advances make the tokenizer a critical upstream component: regardless of whether the downstream generator is autoregressive, masked, multi-scale, or hybrid, unstable VQ training can limit reconstruction fidelity, code utilization, and ultimately generation quality. StableVQ is therefore orthogonal to generator design and aims to provide a more reliable discrete representation substrate for token-based image generation.

\section{Limitations}
\label{sup:limitations}

StableVQ focuses on making VQ tokenizer training stable and reliable, providing a foundation on which stronger training objectives and downstream modeling choices can be explored. A natural direction is to study training recipes that better balance reconstruction quality, semantic structure, and generation-friendliness once codebook utilization and optimization stability are no longer the main bottlenecks. Another promising direction is to connect stable visual tokenization with unified generation and understanding objectives, where discrete tokens may need to preserve both low-level fidelity and high-level semantic information. Finally, while our experiments focus on image tokenizers, the same separation-of-concerns perspective may be useful in other domains that rely on vector quantization, such as video, audio, multimodal representation learning, or compression.

\section{Additional Experimental Results}
\label{sup:results}

We provide additional baselines and experimental results for reference.
For reconstruction, Table~\ref{tab:sup_full_recon} supplements the main results with metrics from methods not covered in the main text, such as VQGAN and MaskGIT. We also provide more complete metrics for selected baselines under the same evaluation script in Table~\ref{tab:sup_same_eval}, including PSNR, SSIM, and other reference metrics.

\begin{table}[h]
\caption{Complete reconstruction results on ImageNet $256\times256$. Codebook Size is reported as $n \times d$ (number of codes $\times$ channel dimension). $^\dagger$ denotes a larger encoder--decoder trained for up to 330 epochs. $^\ddagger$ denotes training for 120 epochs.}
\label{tab:sup_full_recon}
\centering
\small
\setlength{\tabcolsep}{6pt}
\renewcommand{\arraystretch}{1.15}
\begin{tabular}{ccccccc}
    \toprule
    Method & Projector & Tokens & Codebook Size ($n \times d$) & rFID$\downarrow$ & LPIPS$\downarrow$ & Usage$\uparrow$ \\
    \midrule

    VQGAN~\cite{vqgan}    & --- & $16\times16$ & $1{,}024 \times 256$   & 7.94 & ---    & 44\%   \\
    VQGAN~\cite{vqgan}    & --- & $16\times16$ & $16{,}384 \times 256$  & 4.98 & 0.2843 & 5.9\%  \\

    SD-VQGAN~\cite{latent_diffusion} & --- & $16\times16$ & $16{,}384 \times 8$    & 5.15 & ---    & ---    \\
    MaskGIT~\cite{maskgit}  & --- & $16\times16$ & $1{,}024 \times 256$   & 2.28 & ---    & ---    \\

    LlamaGen~\cite{llamagen} & --- & $16\times16$ & $16{,}384 \times 8$    & 2.19 & 0.2281 & 97\%   \\
    LlamaGen~\cite{llamagen} & --- & $16\times16$ & $16{,}384 \times 256$  & 9.21 & ---    & 0.29\% \\

    IBQ$^\dagger$~\cite{ibq}  & --- & $16\times16$ & $16{,}384 \times 256$  & 1.37 & 0.2235 & 96\%   \\
    IBQ$^\dagger$~\cite{ibq}  & --- & $16\times16$ & $262{,}144 \times 256$ & 1.00 & 0.2030 & 84\%   \\

    \midrule
    \multicolumn{7}{l}{\textit{Shared-Projection-Based Methods}} \\
    \midrule

    VQGAN-LC~\cite{vqgan-lc} & Linear-1   & $16\times16$ & $16{,}384 \times 8$    & 3.01 & 0.2358 & 99\%  \\
    VQGAN-LC~\cite{vqgan-lc} & Linear-1   & $16\times16$ & $100{,}000 \times 8$   & 2.62 & 0.2212 & 99\%  \\

    SimVQ~\cite{simvq}    & Linear-1   & $16\times16$ & $16{,}384 \times 256$  & 2.89 & 0.2492 & 100\% \\
    SimVQ~\cite{simvq}    & Linear-1   & $16\times16$ & $262{,}144 \times 256$ & 3.16 & 0.2516 & 100\% \\

    FVQ~\cite{fvq}      & ViTBlock-2 & $16\times16$ & $16{,}384 \times 256$  & 1.70 & 0.2176 & 100\% \\
    FVQ~\cite{fvq}      & ViTBlock-2 & $16\times16$ & $262{,}144 \times 256$ & 1.29 & 0.2003 & 100\% \\

    \midrule

    StableVQ & Linear-1 & $16\times16$ & $16{,}384 \times 256$  & 1.22 & 0.2235 & 100\% \\
    
    StableVQ & Linear-1 & $16\times16$ & $262{,}144 \times 256$  & 1.05 & 0.1947 & 100\% \\
    
    \phantom{$^\ddagger$}StableVQ$^\ddagger$ & Linear-1 & $16\times16$ & $16{,}384 \times 256$  & \textbf{1.13} & \textbf{0.2134} & 100\% \\
    
    \phantom{$^\ddagger$}StableVQ$^\ddagger$ & Linear-1 & $16\times16$ & $262{,}144 \times 256$  & \textbf{0.92} & \textbf{0.1893} & 100\% \\

    \bottomrule
\end{tabular}
\end{table}

\begin{table}[h]
\caption{Additional reconstruction metrics under the same evaluation script on ImageNet $256\times256$.}
\label{tab:sup_same_eval}
\centering
\small
\setlength{\tabcolsep}{6pt}
\renewcommand{\arraystretch}{1.15}
\begin{tabular}{cccccccc}
    \toprule
    Method & Codebook Size ($n \times d$) & Epochs & rFID$\downarrow$ & PSNR$\uparrow$ & SSIM$\uparrow$ & LPIPS$\downarrow$ & Usage$\uparrow$ \\
    \midrule
    SimVQ~\cite{simvq} & $16{,}384 \times 256$  & 40 & 2.89 & 21.36 & 0.5551 & 0.2492 & 100\% \\
    SimVQ~\cite{simvq} & $262{,}144 \times 256$ & 40 & 3.16 & 21.31 & 0.5527 & 0.2516 & 100\% \\
    FVQ~\cite{fvq}     & $16{,}384 \times 256$  & 40 & 1.70 & 21.85 & 0.5894 & 0.2176 & 100\% \\
    FVQ~\cite{fvq}     & $262{,}144 \times 256$ & 40 & 1.29 & 22.50 & 0.6146 & 0.2003 & 100\% \\
    FVQ~\cite{fvq}     & $16{,}384 \times 256$  & 120 & 1.46 & 21.91 & 0.5951 & 0.2156 & 100\% \\
    FVQ~\cite{fvq}     & $262{,}144 \times 256$ & 120 & 1.07 & 22.55 & 0.6244 & 0.1965 & 100\% \\
    \midrule
    StableVQ & $16{,}384 \times 256$  & 40 & 1.22 & 21.84 & 0.5816 & 0.2235 & 100\% \\
    StableVQ & $262{,}144 \times 256$ & 40 & 1.05 & 22.77 & 0.6277 & 0.1947 & 100\% \\
    StableVQ & $16{,}384 \times 256$  & 120 & \textbf{1.13} & \textbf{22.01} & \textbf{0.5965} & \textbf{0.2134} & 100\% \\
    StableVQ & $262{,}144 \times 256$ & 120 & \textbf{0.92} & \textbf{22.81} & \textbf{0.6345} & \textbf{0.1893} & 100\% \\
    \bottomrule
\end{tabular}
\end{table}

For downstream generation, we train class-conditional autoregressive transformers following IBQ~\cite{ibq} on top of StableVQ tokens and evaluate on ImageNet $256\times256$ using FID, IS, precision, and recall. As shown in Table~\ref{tab:sup_full_gen}, StableVQ achieves competitive generation quality, validating that improved reconstruction quality translates to gains in downstream generation.

\begin{table}[h]
\caption{Class-conditional image generation results on ImageNet $256\times256$.}
\label{tab:sup_full_gen}
\centering
\small
\setlength{\tabcolsep}{6pt}
\renewcommand{\arraystretch}{1.15}
\begin{tabular}{ccccccccc}
    \toprule
    Type & Tokenizer & Generator & Param. & FID$\downarrow$ & IS$\uparrow$ & Pre.$\uparrow$ & Rec.$\uparrow$ \\
    \midrule
    Diff.  & SD-VAE~\cite{latent_diffusion}   & DiT-L/2~\cite{dit}       & 458M & 5.02  & 167.2 & 0.75 & 0.57 \\
    Diff.  & SD-VAE~\cite{latent_diffusion}   & DiT-XL/2~\cite{dit}      & 675M & 2.27  & 278.2 & 0.83 & 0.57 \\
    Mask.  & VQGAN~\cite{vqgan}    & MaskGIT~\cite{maskgit}       & 227M & 6.18  & 182.1 & 0.80 & 0.51 \\
    VAR    & VAR~\cite{var}      & VAR-d16~\cite{var}       & 310M & 3.30  & 274.4 & 0.84 & 0.51 \\
    VAR    & VAR~\cite{var}      & VAR-d20~\cite{var}       & 600M & 2.57  & 302.6 & 0.83 & 0.56 \\
    \midrule
    AR     & LlamaGen~\cite{llamagen} & LlamaGen-L~\cite{llamagen}    & 343M & 3.80  & 248.3 & 0.83 & 0.51 \\
    AR     & LlamaGen~\cite{llamagen} & LlamaGen-XL~\cite{llamagen}   & 775M & 3.39  & 227.1 & 0.81 & 0.54 \\
    AR     & FVQ~\cite{fvq}      & LlamaGen-L~\cite{llamagen}    & 343M & 2.39  & 276.6 & 0.84 & 0.56 \\
    AR     & FVQ~\cite{fvq}      & LlamaGen-XL~\cite{llamagen}   & 775M & 2.07  & 287.0 & 0.83 & 0.58 \\
    AR     & IBQ~\cite{ibq}      & IBQ-B~\cite{ibq}         & 342M & 2.88  & 254.7 & 0.84 & 0.51 \\
    AR     & IBQ~\cite{ibq}      & IBQ-L~\cite{ibq}         & 649M & 2.45  & 267.5 & 0.83 & 0.52 \\
    \midrule
    AR     & StableVQ & IBQ-B~\cite{ibq}   & 342M & 2.35 & 256.0 & 0.82 & 0.58 \\
    AR     & StableVQ & IBQ-L~\cite{ibq}   & 649M & 2.18 & 250.4 & 0.82 & 0.59 \\
    \bottomrule
\end{tabular}
\end{table}

\section{Additional Analysis Experiments}
\label{sup:analysis}

\begin{figure}[h]
    \centering
    \includegraphics[width=\linewidth]{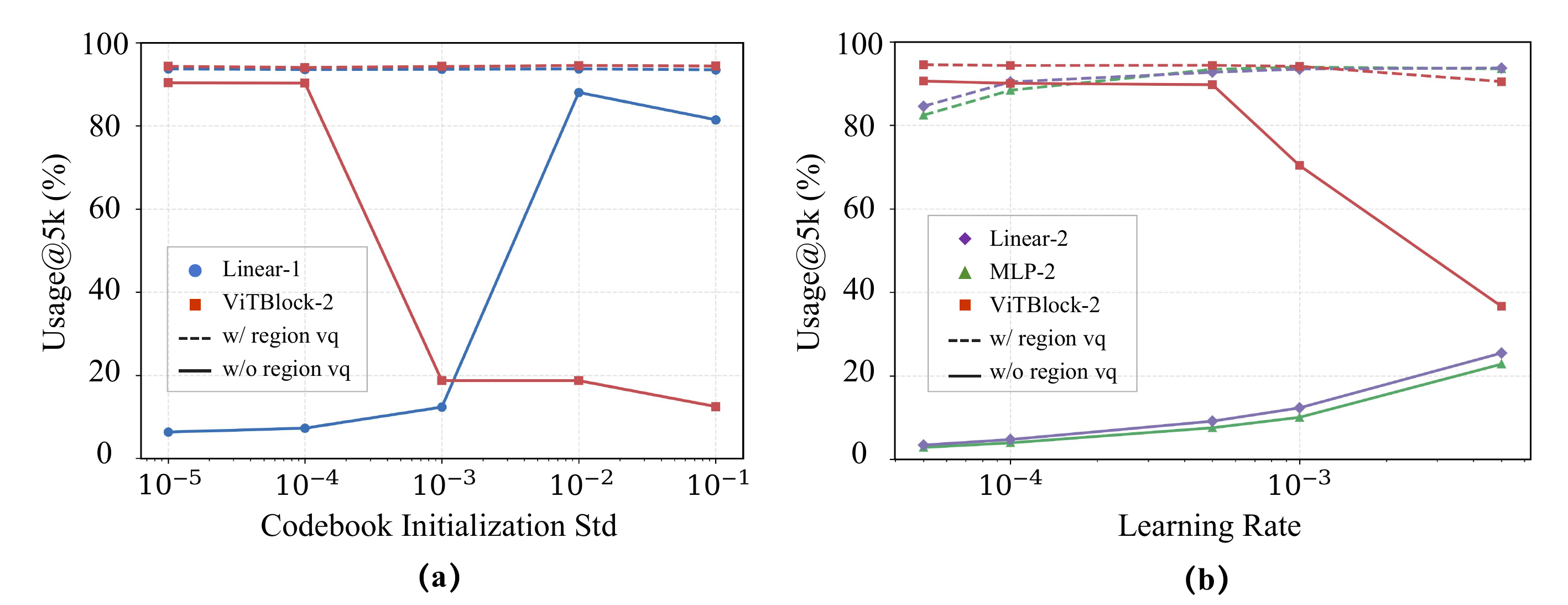}
    \vspace{-10pt}
    \caption{
        Additional analysis experiments. (a) Usage@5k under different Gaussian codebook initialization scales. (b) Usage@5k under different learning rates.
    }
    \label{fig:additional_analysis}
    \vspace{-10pt}
\end{figure}

\subsection{Effect of Codebook Initialization}
\label{sup:init_scale}

Codebook initialization has long been an important practical trick in conventional VQ training, because an inappropriate initialization scale can lead to poor codebook utilization. A common strategy is to initialize the Codebook within a small numerical range, which often helps utilization increase more stably. However, this trick is not always applicable: for example, when an $\ell_2$ normalization constraint is imposed, the effective code distribution can no longer be controlled simply by shrinking the raw initialization range. Moreover, an overly small initialization scale may also prolong the usage-recovery phase, reducing training efficiency.

To study how initialization affects codebook utilization across configurations, we follow the pilot setting in Section~\ref{sec:region_vq}: the Encoder is frozen, and only the Codebook is optimized to fit a fixed target distribution. We vary the Gaussian initialization scale of the Codebook and report codebook utilization at 5k steps, computed within a window of 65,536 tokens. As shown in Figure~\ref{fig:additional_analysis}(a), when the initialization scale is too small, the linear shared projector used in SimVQ-style settings recovers utilization slowly; when the initialization scale becomes larger, the ViT-based shared projector used in FVQ-style settings faces a clear dead-code risk. Replacing the standard VQ loss with Region VQ Loss substantially improves both cases, allowing utilization to approach full usage at 5k steps across initialization scales. This verifies that StableVQ reduces the dependence of VQ training on codebook initialization design, and also suggests robustness to scale divergence during training.

\subsection{Effect of Shared Projector}
\label{sup:shared_projector}

For methods that rely on a shared projector to improve codebook utilization, projector design has recently become an important practical consideration. When the projector is the simplest linear layer, its limited learning capacity can make it difficult for the Codebook to quickly track the encoder output distribution, often leading to slow utilization growth and more frequent scale-divergence events during training. In contrast, using a more expressive ViT block can improve tracking capacity, but introduces additional architecture-specific design cost and constraints on codebook size.

To examine how shared projector design affects utilization, we again follow the frozen-Encoder pilot setting and measure codebook utilization at 5k steps under different learning rates, using the same 65,536-token window. The Codebook is initialized with a small Gaussian standard deviation of $10^{-4}$, so the results more directly reflect projector learning capacity while reducing the influence of dead-code effects. As shown in Figure~\ref{fig:additional_analysis}(b), neither a two-layer linear projector nor a two-layer MLP can raise utilization to a high level within 5k steps, even when the learning rate is increased to $5\times10^{-3}$. The two-layer ViT block performs better at moderate learning rates, but its utilization drops sharply when the learning rate becomes too large. With Region VQ Loss, however, different projector structures all achieve strong utilization within 5k steps, and their usable learning-rate range is substantially widened. This suggests that the Codebook benefits from an independent learning-rate schedule for tracking the encoder distribution, but also shows that changing projector capacity alone is insufficient for stable and rapid utilization growth. The more critical factor is to provide the Codebook with a principled learning objective, which allows StableVQ to improve utilization across shared-projector designs and reduces the dependence of robust VQ training on carefully engineered projector structures.

\subsection{Computational and Memory Overhead}
\label{sup:overhead}

Table~\ref{tab:sup_overhead} reports matched baseline and StableVQ profiles with 16K and 262K codebooks on the same machine. The Decoupled Schedule adds no forward or backward operation, and the measured time difference of Dynamic STE is within noise. The cost of Region VQ Loss decreases as codebook utilization increases because fewer inactive codes require propagated targets. All components are training-only and leave tokenizer inference unchanged.

\begin{table*}[h]
\caption{Training-time and memory overhead of StableVQ components. Parentheses denote changes from the corresponding baseline, and all memory values are per device.}
\label{tab:sup_overhead}
\centering
\scriptsize
\setlength{\tabcolsep}{3pt}
\renewcommand{\arraystretch}{1.15}
\begin{tabular*}{\textwidth}{@{\extracolsep{\fill}}lccccccc@{}}
    \toprule
    & \multicolumn{5}{c}{Training Cost} & \multicolumn{2}{c}{Memory Usage} \\
    \cmidrule(lr){2-6}\cmidrule(lr){7-8}
    Codebook
    & Baseline
    & Dynamic STE
    & Region VQ @30\%
    & Region VQ @50\%
    & Region VQ @90\%
    & Baseline
    & Region VQ \\
    & (ms) & (ms) & (ms, $\Delta$) & (ms, $\Delta$) & (ms, $\Delta$) & (GiB) & (GiB, $\Delta$) \\
    \midrule
    16K
    & 438.1
    & 437.2
    & 441.2 ($+0.71\%$)
    & 440.4 ($+0.51\%$)
    & 440.2 ($+0.46\%$)
    & 34.06
    & 34.07 ($+0.04\%$) \\
    262K
    & 462.2
    & 462.7
    & 472.2 ($+2.16\%$)
    & 469.2 ($+1.50\%$)
    & 466.2 ($+0.85\%$)
    & 35.12
    & 35.55 ($+1.24\%$) \\
    \bottomrule
\end{tabular*}
\end{table*}

\subsection{Comparison with Prior STE Corrections}
\label{sup:ste_corrections}

VQ-STE++~\cite{straighten} employs Alternating Optimization, which is similar in motivation to Dynamic STE: both aim to suppress unreliable task gradients when quantization error is large. However, VQ-STE++ relies on alternating inner and outer updates, introducing several sensitive hyperparameters and additional training overhead. It removes the encoder commitment loss and assumes an initially aligned code--token distribution established through $k$-means initialization, making optimization sensitive when this alignment is absent.

The Rotation Trick~\cite{rotationtrick} transforms the encoder gradient through a rotation matrix $R$ and a rescaling factor $\lVert q\rVert/\lVert z\rVert$. For rotation, $R$ redirects the backward gradient according to the angle between token $z$ and code $q$. Although elegant, it provides no rigorous guarantee of a more accurate gradient and may instead redirect the gradient away from the desired direction. For rescaling, the factor suppresses the encoder gradient when $\lVert q\rVert<\lVert z\rVert$, but amplifies it when $\lVert q\rVert>\lVert z\rVert$, even for distant token--code pairs. Dynamic STE instead uses relative within-batch quantization distances to attenuate unreliable gradients and never amplify them. Moreover, under $\ell_2$ normalization, the rescaling factor becomes one and loses its attenuation effect, whereas Dynamic STE remains active.

We conduct a controlled 15-epoch experiment with Region VQ and all other experimental settings fixed. As shown in Table~\ref{tab:sup_ste_corrections}, VQ-STE++ consistently exhibits a pronounced collapse in codebook utilization without the joint use of $k$-means initialization to pre-align the token--code distributions and a norm bottleneck. In contrast, both the Rotation Trick and Dynamic STE achieve full codebook utilization with the support of Region VQ, while Dynamic STE yields substantially better reconstruction performance across metrics.

\begin{table*}[h]
\caption{Comparison of STE corrections in training stability and reconstruction quality.}
\label{tab:sup_ste_corrections}
\centering
\small
\setlength{\tabcolsep}{6pt}
\renewcommand{\arraystretch}{1.15}
\begin{tabular*}{\textwidth}{@{\extracolsep{\fill}}lccccc@{}}
    \toprule
    Method & Training Status & Usage$\uparrow$ & rFID$\downarrow$ & PSNR$\uparrow$ & LPIPS$\downarrow$ \\
    \midrule
    VQ-STE++~\cite{straighten} & Collapsed & -- & -- & -- & -- \\
    Rotation Trick~\cite{rotationtrick} & Completed & 100.00\% & 3.6437 & 20.4024 & 0.2568 \\
    Dynamic STE (ours) & Completed & 100.00\% & \textbf{2.7942} & \textbf{21.1882} & \textbf{0.2359} \\
    \bottomrule
\end{tabular*}
\end{table*}

\subsection{Comparison with Explicit Distribution-Alignment Methods}
\label{sup:distribution_alignment}

Wasserstein VQ~\cite{wassersteinvq} and MMD VQ~\cite{vqtransplant} formulate global distribution matching as a training loss, using Gaussian moments and kernel statistics, respectively. Region VQ Loss instead addresses the asymmetric supervision of standard VQ by assigning persistently inactive codes explicit, local targets propagated from statistically supported active regions. It therefore requires no parametric assumption about the token distribution and avoids the costly pairwise kernel computation of MMD VQ.

We evaluate these methods on the non-Gaussian mixture benchmark introduced by VQ-Transplant~\cite{vqtransplant}. At $\zeta=0$, the target distribution reduces to a single Gaussian; increasing $\zeta$ separates the two mixture modes and progressively strengthens its non-Gaussian structure. Following its setting, we use 16,384 codes of dimension 8, sample 20K tokens per step, and report codebook utilization at 10K steps. We quote the Wasserstein VQ and MMD VQ utilization results from Table~13 of VQ-Transplant and evaluate Region VQ on the same $\zeta$ grid with a linear shared projector. We additionally measure the per-step training time of all methods under the same environment. To mitigate under-coverage caused by heavily overlapping recipient sets in this synthetic benchmark, we automatically enlarge the propagation quotas when recipient collisions are severe. 

Table~\ref{tab:sup_distribution_alignment} reports the resulting codebook utilization across different $\zeta$ values and the corresponding per-step training time. As the target distribution becomes strongly non-Gaussian, Wasserstein VQ and MMD VQ fall to 34.8\% and 75.6\% utilization at $\zeta=4$, respectively, whereas Region VQ maintains 99.4\%. Region VQ also remains close to Wasserstein VQ in training time and is $34.2\times$ faster than MMD VQ. These results demonstrate that Region VQ combines robust codebook utilization across different distribution relationships with low training overhead.

\begin{table}[h]
\caption{Codebook utilization and training efficiency on the non-Gaussian distribution-fitting benchmark. Utilization is measured at 10K steps. Relative time is normalized to Wasserstein VQ.}
\label{tab:sup_distribution_alignment}
\centering
\small
\setlength{\tabcolsep}{5pt}
\renewcommand{\arraystretch}{1.15}
\begin{tabular}{lccccccc}
    \toprule
    Method & $\zeta=0$ & $\zeta=1$ & $\zeta=2$ & $\zeta=3$ & $\zeta=4$ & Time (ms) & Relative Time \\
    \midrule
    Wasserstein VQ~\cite{wassersteinvq} & 99.9\% & 97.0\% & 62.7\% & 44.8\% & 34.8\% & 6.98 & $1.00\times$ \\
    MMD VQ~\cite{vqtransplant} & 99.9\% & 99.8\% & 92.5\% & 85.7\% & 75.6\% & 295.57 & $42.36\times$ \\
    Region VQ (ours) & 99.8\% & 99.3\% & 93.2\% & 99.4\% & 99.4\% & 8.64 & $1.24\times$ \\
    \bottomrule
\end{tabular}
\end{table}

\section{Robustness Test Illustration}
\label{sup:robustness_test}

The robustness test is designed to evaluate whether a VQ training method can recover high codebook utilization when the relationship between the Codebook distribution and the token distribution changes. This setting complements the final validation usage reported in the main reconstruction table: a method may eventually report high utilization under its standard configuration, yet still recover slowly or fail when the codebook-token relationship becomes less favorable. We control this distributional relationship through different codebook initializations.

We construct different codebook-token distribution relationships by varying the Gaussian initialization scale of the codebook vector base while keeping the data, architecture, optimizer, learning-rate schedule, and training budget fixed. Changing this scale alters the initial geometry between the projected code vectors and the encoder token distribution: small scales place code vectors in a compact region, whereas larger scales spread the code distribution over a broader region relative to the token distribution. This provides a controlled way to test codebook usage recovery under multiple mismatch settings without changing the input data or reconstruction objective.

For each method and each mismatch setting, we train the tokenizer for the same early-stage budget and record codebook usage throughout training. Usage is computed within a window of 65,536 tokens, so the reported values are slightly lower than utilization measured over the full validation set. Figure~\ref{fig:usage_recovery} visualizes these trajectories as heatmaps: each row corresponds to one mismatch setting, and brighter colors indicate higher codebook usage. A stable method should recover high usage quickly across most rows rather than depending on a narrow range of favorable initial relationships.

We summarize the heatmaps with UR-AUC, which denotes Usage Recovery AUC. Let $U_m(t)$ be the codebook usage percentage at training step $t$ under mismatch setting $m$. We compute the normalized area under each usage curve and average over the tested mismatch settings:
\begin{equation}
    \mathrm{UR\text{-}AUC}
    =
    \frac{1}{|\mathcal{M}|}
    \sum_{m\in\mathcal{M}}
    \frac{1}{T - t_0}
    \int_{t_0}^{T} U_m(t)\,dt .
\end{equation}
In practice, we use the logged usage values and compute the integral with the trapezoidal rule. Higher UR-AUC indicates faster and more consistent recovery of codebook utilization under codebook-token distribution mismatch. The error bars for UR-AUC in Table~\ref{tab:main_recon} are computed as the sample mean and sample standard deviation over three runs with different random seeds, where each run includes the same set of initialization-induced mismatch settings described above.

\section{Region VQ Algorithm}
\label{sup:algo}

Algorithm~\ref{alg:region_vq} summarizes the Region VQ procedure in one training step. It follows the design in Section~\ref{sec:region_vq}: current assignments determine active codes and their targets, a FIFO queue identifies persistently inactive codes, and targets from sufficiently reliable active codes are propagated to nearby inactive ones before computing the codebook loss.

\begin{algorithm}[h]
\caption{Region VQ in one training step}
\label{alg:region_vq}
\small
\begin{algorithmic}[1]
\Require Token features $\mathbf{Z}=\{\mathbf{z}_u\}_{u=1}^{N}$, projected codebook $\mathbf{E}=\{\mathbf{e}_k\}_{k=1}^{K}$, current assignments $a_u \in [K]$, FIFO queue $Q$ of length $W$
\Ensure Code targets $\{\mathbf{t}_k\}_{k=1}^{K}$ and effective code set $\mathcal{M}_t$ for codebook loss
\State Append current assignment indices $\{a_u\}_{u=1}^{N}$ to $Q$
\State Let $\mathcal{S}_t \gets \{a_u\}_{u=1}^{N}$ be the currently active codes
\State Let $\mathcal{A}_t \gets \bigcup Q$ be the codes active within the FIFO window
\State Let $\mathcal{N}_t \gets [K] \setminus \mathcal{A}_t$ be the persistently inactive codes
\State Initialize $\mathbf{t}_k \gets \mathbf{e}_k$ for all $k \in [K]$ \Comment{Self-target by default}
\State Initialize effective code set $\mathcal{M}_t \gets \emptyset$
\For{each $k \in \mathcal{S}_t$}
    \State $\mathcal{U}_k \gets \{u : a_u = k\}$, \quad $n_k \gets |\mathcal{U}_k|$
    \State $\mathbf{t}_k \gets \frac{1}{n_k} \sum_{u \in \mathcal{U}_k} \mathbf{z}_u$
    \State Add $k$ to $\mathcal{M}_t$
\EndFor
\State Let $\mathcal{S}_t^{+} \gets \{k \in \mathcal{S}_t : n_k > 1\}$
\For{each $k \in \mathcal{S}_t^{+}$}
    \State $q_k \gets \left\lceil n_k |\mathcal{N}_t| \,/\, \sum_{\ell \in \mathcal{S}_t^{+}} n_\ell \right\rceil$
    \State Select the $q_k$ nearest codes to $k$ from $\mathcal{N}_t$ in projected space and denote them by $\mathcal{R}_k$
\EndFor
\For{each $j \in \mathcal{N}_t$ that is selected by at least one active code}
    \State Let $\mathcal{K}_j \gets \{k \in \mathcal{S}_t^{+} : j \in \mathcal{R}_k\}$
    \State $\mathbf{t}_j \gets \frac{1}{|\mathcal{K}_j|} \sum_{k \in \mathcal{K}_j} \mathbf{t}_k$
    \State Add $j$ to $\mathcal{M}_t$
\EndFor
\State Compute codebook loss only on effective codes:
$\mathcal{L}_{\mathrm{code}} = \frac{1}{|\mathcal{M}_t|} \sum_{k \in \mathcal{M}_t} \|\mathbf{e}_k - \mathrm{sg}[\mathbf{t}_k]\|_2^2$

\end{algorithmic}
\end{algorithm}

\section{Experimental Details}
\label{sup:details}

\label{sup:main_config}

We provide the detailed configurations used for reconstruction training in Table~\ref{tab:sup_main_config}. For the generation experiments in Appendix~\ref{sup:results}, we follow the corresponding IBQ training setting and train for approximately 350 epochs. Other experiments in this paper describe their key differences from the standard settings in their respective contexts.

\begin{table}[h]
\caption{Main experiment configurations.}
\label{tab:sup_main_config}
\centering
\small
\setlength{\tabcolsep}{7pt}
\renewcommand{\arraystretch}{1.15}
\begin{tabular}{lccc}
    \toprule
    Config & SimVQ & FVQ & StableVQ \\
    \midrule
    Base Batch Size & 128 & 128 & 128 \\
    Training Epochs & 40 & 40 / 120 & 40 / 120 \\
    Optimizer & Adam & Adam & Adam \\
    Optimizer Parameters & $\beta_1=0.9,\beta_2=0.95$ & $\beta_1=0.9,\beta_2=0.95$ & $\beta_1=0.9,\beta_2=0.95$ \\
    Base Learning Rate & $1\mathrm{e}{-4}$ & $1\mathrm{e}{-4}$ & $1\mathrm{e}{-4}$ \\
    Codebook Init Type & Gaussian & Uniform & Uniform \\
    \midrule
    LR Warmup & 0\% & 10\%, $0.005\times$ to $1\times$ & 10\%, $0.005\times$ to $1\times$ \\
    LR Plateau & 100\%, $1\times$ & 27\%, $1\times$ & 27\%, $1\times$ \\
    LR Annealing & 0\% & 63\%, $1\times$ to $0.01\times$ & 63\%, $1\times$ to $0.01\times$ \\
    Codebook LR & Same as LR & Same as LR & Constant $1\mathrm{e}{-3}$ \\
    \bottomrule
\end{tabular}
\end{table}

\section{Visualization Results}
\label{sup:visual}

\begin{figure}[h]
    \centering
    \includegraphics[width=\linewidth]{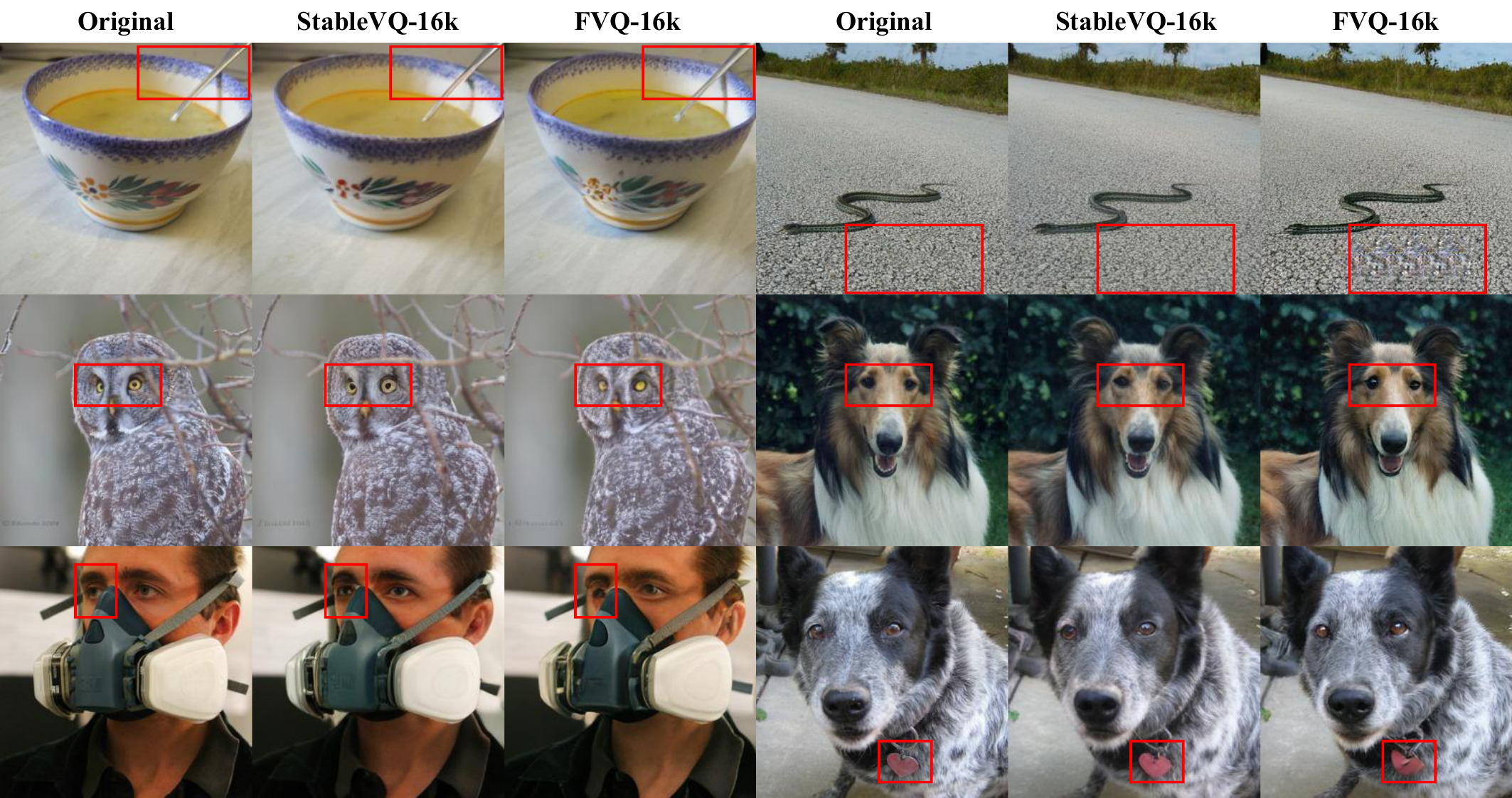}
    \vspace{-10pt}
    \caption{Qualitative reconstruction comparison with a 16k codebook.}
    \label{fig:sup_recon_16k}
\end{figure}

\begin{figure}[h]
    \centering
    \includegraphics[width=\linewidth]{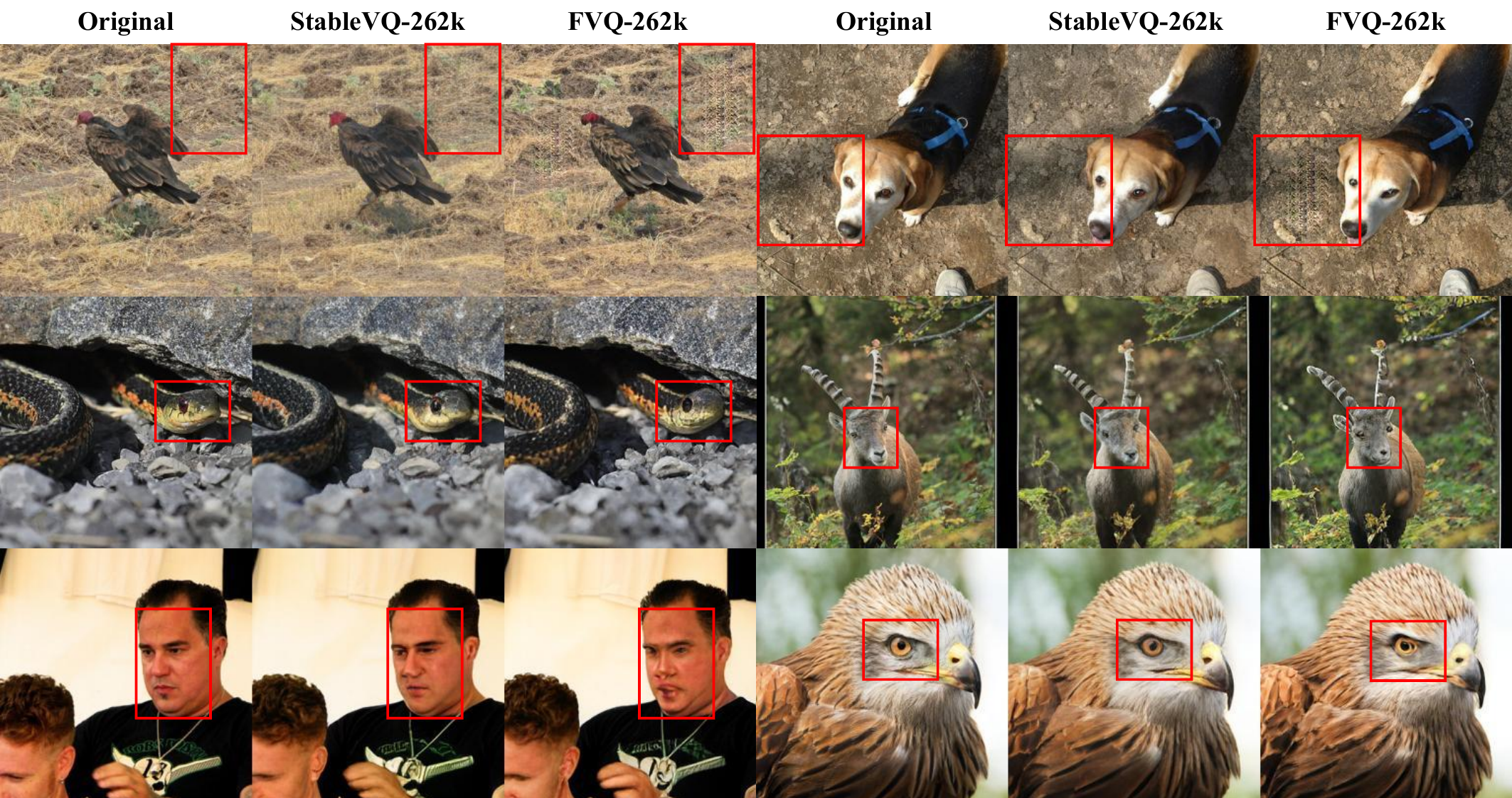}
    \vspace{-10pt}
    \caption{Qualitative reconstruction comparison with a 262k codebook.}
    \label{fig:sup_recon_262k}
\end{figure}

\begin{figure}[h]
    \centering
    \includegraphics[width=\linewidth]{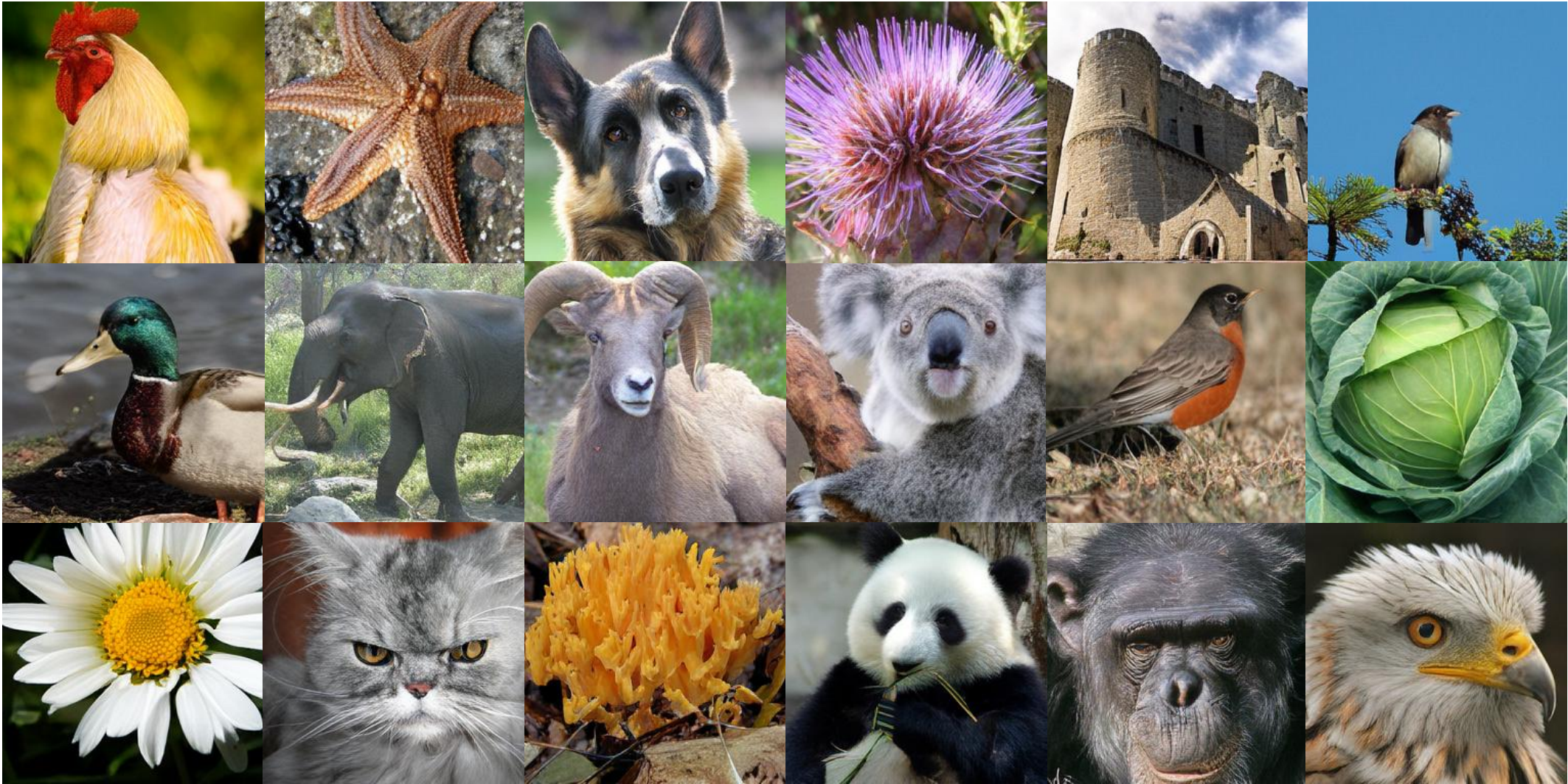}
    \caption{Qualitative samples from the downstream class-conditional image generation task.}
    \label{fig:sup_generation}
\end{figure}

We provide qualitative visualization results for reconstruction and downstream generation.
Figure~\ref{fig:sup_recon_16k} compares reconstruction results with a 16k-code codebook, Figure~\ref{fig:sup_recon_262k} compares reconstruction results with a 262k-code codebook, and Figure~\ref{fig:sup_generation} shows samples from the downstream generation task.

\vspace{-10pt}

\end{document}